\documentclass[12pt]{article}

\usepackage{url}
\usepackage[T1]{fontenc}
\usepackage{amsmath}
\usepackage{newtxtext,newtxmath} % TeX Gyre Termes, as in the supplied PDF
\usepackage[letterpaper,margin=1in]{geometry}
\usepackage{setspace}
\usepackage{graphicx}
\usepackage{placeins}        % keep all supplied figures within Results
\graphicspath{{figures/}}

\usepackage[labelfont=normalfont,labelsep=colon,font=normalsize,
            justification=justified,singlelinecheck=false]{caption}
\newenvironment{sciabstract}{%
  \begin{quote}\bfseries}{\end{quote}}

\makeatletter
\renewcommand\@cite[2]{(\textit{#1\if@tempswa , #2\fi})}
\renewcommand\@biblabel[1]{#1.}
\makeatother

\newcommand{\figimg}[3][0.56\textheight]{%
  \IfFileExists{figures/#3}%
    {\includegraphics[width=#2,height=#1,keepaspectratio]{figures/#3}}%
    {\fbox{\parbox[c][0.25\textheight][c]{\dimexpr#2-2\fboxsep-2\fboxrule\relax}{\centering\small Missing figure file:\\\texttt{\detokenize{#3}}}}}}

\title{\bfseries Generated storm lifecycles fill the rare-event tail of short rainfall records}
\author{
Valerie Tsao$^{1\ast}$, Nathaniel Chaney$^{1}$, Manolis Veveakis$^{1}$, Marta Zaniolo$^{1}$\\
\small{$^{1}$Department of Civil and Environmental Engineering, Duke University, Durham, NC 27708, USA.}\\
\small{$^\ast$Corresponding author. Email: valerie.tsao@duke.edu}
}

\date{}

\begin{document}
\maketitle

%\begin{sciabstract}
%\indent Flood defenses must withstand storms that short rainfall records rarely contain, leaving planners to extrapolate from a handful of extremes. We present StormGen, a generative model that turns an observed storm’s first hours into complete hourly rainfall lifecycles in minutes. Trained on a decade of southeast Texas rainfall, StormGen produced close counterparts for 45\% of the rarest storms, comparable to held-out observations and against 3\% for stochastic storm transposition and kept storms coherent for 24 hours where an AI nowcast faded after six. From a single hour of rainfall, it generated Harvey-scale catchment totals for a hurricane withheld from training. Because timing alone can more than double peak flows from storms of equal depth, complete scenarios can stress-test flood protection against extremes beyond the record.
%\end{sciabstract}

\begin{sciabstract}
\indent Historically rare "one-in-a-generation" storms are occurring with unprecedented frequency, causing catastrophic damage and mounting casualties, and exposing the vulnerability of infrastructure designed around short observational records. To overcome the limitations of sparse data, we present StormGen, a generative machine-learning model that transforms an observed storm's initial hours into complete, physically plausible hourly rainfall lifecycles in minutes. Evaluated on a decade of southeast Texas rainfall, StormGen produced close counterparts for 45\% of the most severe 2\% of storms—compared to just 3\% for stochastic storm transposition—and maintained spatial coherence over full 24-hour sequences. Crucially, from a single hour of initial rainfall, the model successfully reconstructed the specific lifecycle, spatial pattern, and extreme catchment totals of Hurricane Harvey itself, despite the event being entirely absent in training. Because storm timing alone can more than double peak flows, these complete generated scenarios provide a vital mechanism to stress-test flood protections against unprecedented extremes.
\end{sciabstract}

\section{Introduction}

Storms that deliver extreme rainfall are among the most destructive natural hazards on Earth. Between 1970 and 2021, weather-, climate-, and water-related disasters killed over 2 million people and caused around US\$4.3 trillion in economic losses, and storms (tropical cyclones in particular) were the leading source of those losses \cite{wmo2023atlas}. Roughly 1.81 billion people live where a 100-year flood would reach them, 89\% of them in low- and middle-income countries \cite{Rentschler2022_FloodExposure}, and the most extreme rainfall that drives such floods is projected to intensify by about 7\% for every additional Celsius of warming \cite{ipcc2021ch11}. These figures point to a global problem, but not an equal one. Extreme rainfall threatens nearly a quarter of the world’s population, but its heaviest toll falls on poorer countries, and a warming climate is exacerbating this. In the United States alone, 403 disasters since 1980 each caused at least \$1 billion in damage (about \$2.9 trillion in total), and their frequency has risen from 9 per year (1980 to 2023) to 23 per year (2020 to 2024) \cite{noaancei2025billions}. Defenses against these disasters, from the size of a storm drain to the decision of where to build, rest on estimates of how large and how frequent the worst storms can be.

Engineers design against benchmarks such as the 100-year storm (i.e., the rainfall depth with a 1\% chance of being exceeded in any given year), but reliably estimating such rare events requires long rainfall records that capture a sufficiently broad range of extremes. When a record understates the worst storm a place can produce, the error propagates into every decision built on it. Houston, the largest city in our study region and a comparatively well-monitored one, shows that such errors are not confined to data-poor places. When NOAA updated its Texas estimates in 2018 to include Hurricane Harvey, the 100-year, 24-hour estimate around Houston increased by more than 3 inches \cite{Perica2018Texas}, comparable to the 16-19 inches previously mapped for the 500-year, 1-day storm \cite{ Asquith2004Atlas}. Infrastructure built to the old standard was, in other words, designed for storms far smaller than the region can produce.

Estimating such extremes is hampered by several challenges. First, the events that matter most are, by definition, the least observed.  Long rain-gauge records measure rainfall at individual locations, potentially missing nearby extremes and leaving the spatial structure of storms unresolved. Gridded  hourly, kilometer-scale rainfall analyses can capture spatial structure but span about two decades at most (e.g. Stage IV since 2022 \cite{ncep_stage4_data}); longer km-scale records such as AORC and CONUS404 are model-based reconstructions. Second, flood impacts depend not only on peak rainfall but on a storm’s duration, extent, and movement, combinations that short records sample even more sparsely. Third, records are shortest where exposure is highest, as observing networks in many low- and middle-income countries fall short of international standards \cite{Su2026}. This means that the communities most exposed to flooding are also the least able to anticipate it or recover from it, and the institutions that serve them must plan with the least information. The broader social implications of this disparity are multifaceted, but expanding data availability remains a fundamental prerequisite for developing effective adaptation and mitigation strategies within disproportionately burdened frontline communities. Doing so requires directly overcoming the relative paucity of crisis scenarios captured in existing records, and bridging this data gap provides a pathway to better characterizing extreme events, and, in turn, to more robust risk models. The same shortage of observed extremes limits risk estimates for heat waves \cite{Thompson2023}, droughts \cite{ault2016relative}, and wildfires \cite{Squire2021}, which makes extending short records a problem for disaster science at large.

To look beyond the record, researchers have relied on numerical weather prediction (NWP) models to derive insights into modeling extreme events, as well as Extreme Value Theory (EVT), which provides a statistical method to simulate rare scenarios by fitting mathematical curves to the largest observed values \cite{RADFAR2023100575, boulaguiem2022modeling}. Stochastic storm transposition (SST), widely used in flood hydrology, enlarges a local record by borrowing storms observed elsewhere in the surrounding region and relocating them over the site of interest \cite{WRIGHT201734}. Nevertheless, it cannot create combinations of duration, intensity, and extent that regional archive simply does not have data for.
Generative machine-learning models could, in principle, produce those missing storms. Models of this kind (the same technology underlying AI image generators) have recently shown considerable success in applications of weather forecasting, precipitation nowcasting (i.e., predicting rainfall over the next few hours), and climate downscaling (i.e., adding local detail to course climate-model output) \cite{Nai2025, ravuri2021dgmr, ignacio2025dynamical}. These models can learn spatial and temporal patterns from data without being explicitly given the physical equations that govern them. Trained on regional rainfall sequences, they could therefore learn how storms grow, move, and decay, enabling them to generate alternative storms consistent with regional rainfall characteristics even when those particular sequences have never been observed. Whether these generated storms add information to estimates of rare events has, to our knowledge, not been tested. Should they do so, generated storms could supplement sparse observational records, allowing decision-makers to evaluate mitigation investments and adaptation policies against a broader range of plausible extremes than the historical record alone affords. Moreover, because the shortest records tend to belong to the most exposed communities, such a capability could help narrow an asymmetry in hazard information that at present mirror, and reinforces, existing disparities in wealth and adaptive capacity.

Here, we present StormGen, a generative model that augments short rainfall records with alternative storm lifecycles whose physical characteristics are plausible for the region, expanding the range of rainfall scenarios available for risk analysis. Unlike stochastic storm transposition, which resamples and relocates observed events, StormGen generates new rainfall evolutions that are not restricted to those already in the historical catalog. It also moves beyond short-range precipitation nowcasting by generating controllable alternatives spanning an entire storm lifecycle. Conditioned on an observed storm’s initial hours and user-specified storm characteristics, it generates complete sequences of hourly rainfall in minutes. Using a decade of hourly rainfall over southeast Texas, we (i) test whether generated storms reproduce observed ones, including rare long, intense storms; (ii) compare their coherence over a full day with a leading AI nowcasting model; (iii) generate the most extreme storms after withholding them from training; and (iv) measure how generated storms change operational exceedance probabilities and rainfall-frequency estimates from sparse records. Results show that StormGen can expand the representation of rare extreme events; it supplied close counterparts for 45\% of the most severe 2\% of reference storms, compared with 3\% for stochastic storm transposition (SST). In the Greens and Halls Bayou catchments, we compared a uniform Atlas~14 design storm, historical HRRR rainfall patterns, and a StormGen-augmented catalog under the same rainfall-depth and infrastructure constraints. Augmenting a catalog of 12 historical storms changed coverage of independently withheld flood responses from 60\% to 76\%, revealing additional combinations of peak discharge and peak timing that were absent from the short historical catalog. Our results shed light on how much generative models can add to short records of rare extremes, provided that generated storms are counted as scenarios rather than years of record, and point to a way of preparing for storms where the impact is most acute.

% The generated record contains storms with characteristics compatible with 45\% of the most severe 2\% of storms, against 3\% for SST. It also maintained coherent rainfall patterns throughout 24-hour storm sequences, whereas rainfall generated by the nowcasting model faded after about six hours. Added to a record of only 12 observed storms, it reduced rainfall-frequency errors by 36\%, the equivalent of 110 additional observed storms.

% =====================================================================
% Sections 1--7 retain the current manuscript structure and prose.
% Figure-dependent passages and captions updated for the six supplied PDFs.
% =====================================================================

\section{From historical records to alternative storm lifecycles}
We developed StormGen using a decade (2015 to 2024) of hourly rainfall over southeast Texas, on a study tile spanning the San Jacinto River basin and much of the Houston metropolitan area (Fig.~\ref{fig:workflow}a). The rainfall maps come from the High-Resolution Rapid Refresh (HRRR), a NOAA product that combines radar, satellite, and ground observations with a weather model to estimate rainfall every hour on a 3-km grid. From the 85,710 hourly fields in the archive, we extracted 389 uninterrupted storm events and divided them into 268 training, 61 validation, and 60 test events. The region receives rain from both tropical systems and thunderstorms, and its flat, heavily developed watersheds can turn a few hours of intense rain into damaging floods, which makes it a demanding testbed. A separate Southern Great Plains domain, never seen during training, was reserved to test whether StormGen could generate realistic storm lifecycles in a new region with overlapping rainfall processes (Fig.~\ref{figS:nowcast}).

StormGen expands rainfall records with physically plausible storm scenarios by taking the initial hours of observed storms and generating alternative evolutions consistent with regional rainfall characteristics. These scenarios broaden the representation of extremes available for risk analysis beyond those captured in the observed record. Each generation begins with six hours of observed rainfall, which we call the seed, and a brief description of the storm to be generated: how long it lasts, how much rain it delivers across the region, the highest rainfall rate it may reach, and, optionally, how much of the region it covers (Fig.~\ref{fig:workflow}, b and c). A neural network first produces a best guess of the full continuation, and a diffusion model then adds realistic variations to all future hours at once, so that the storm evolves consistently rather than drifting from hour to hour. Repeated sampling from one seed therefore yields an ensemble of distinct but plausible alternatives to a single observed storm (Fig.~\ref{fig:workflow}c).

To be useful for risk analysis, generated storms must meet four requirements, which structure the analyses that follow. They must be realistic while capturing plausible, yet rare or still unseen, combinations of duration and intensity that cause the worst floods; they must coherently reproduce a storm’s full lifecycle; they must remain credible for storms more extreme than any seen in training; and they must improve the hazard estimates on which planners rely, from operational exceedance probabilities to rainfall-frequency curves. We benchmark StormGen against SST, and against a leading AI nowcasting model. 

\section{Event statistics beyond transposition}
Realism is a demanding standard for a generative model. A model can reproduce average rainfall statistics while producing storms that no real atmosphere would, and models trained to minimize pixel-wise error tend to smooth rainfall into fields that lower that error with resembling any single storm. We therefore examined StormGen at three levels: individual storms, the full population of 6,000 generates storms, and the rarest storms within it (Materials and Methods).

\subsection{Lifecycle and marginal statistics}
StormGen reconstructs a storm’s full lifecycles from its onset. In a representative January 2023 storm over San Jacinto, for instance, the seed conditioned the model only on the first hours of rain, when average rainfall across the region was about 1mm\, h${-1}$ (Fig.~\ref{fig:workflow}, c and d). Over the next five hours, the observed storm intensified roughly fourfold and spread across about 90\% of the region. The ensemble tracked this growth closely, peaking in the same hour at 4.5 against 5.1 mm\,h${-1}$ observed, and both the observed and generated storms had ended about 11 hours after the seed (Fig.~\ref{fig:workflow}, d and e). This matters because the seed itself contains none of the storm’s peak or decay. Although the generated storms remained slightly more widespread than the observed one as it waned, and their median peak was slightly lower, such divergence is to be expected, since the requested characteristics constrain a storm’s overall duration and intensity rather than its hour-by-hour detail. Each member keeps the observed storm’s overall shape while placing its heaviest rain in different locations and reaching different intensities, with local peak rates of about 33 and 47 mm\,h$^{-1}$ in two members (Fig.~\ref{fig:workflow}, c and f), which is the kind of variability a risk analysis needs to sample.
The same fidelity holds across the full population and in how rain is organized in space. The durations, intensities, speeds, areas, and rainfall totals of the 6,000 generated storms matched those of observed storms (Fig.~\ref{figS:storm_properties_profiles}). Their rainfall fields were also nearly indistinguishable from observed ones at every spatial scale, including the finest scales that hold the sharp, localized cores of heavy rain, where transposed storms lost detail. At the finest resolved scale, StormGen retained 89\% of the observed spectral power (a measure of spatial variability in rainfall intensity), compared with 7\% for transposition (Fig.~\ref{fig:fidelity}b). By the end of a 24-hour window, StormGen’s ensemble error, measured by the continuous ranked probability score (CRPS), was on average about one-third lower than that of transposition (0.29 versus 0.44mm\,h$^{-1}$; Fig.~\ref{fig:fidelity}a). This is consistent with reports that diffusion-based precipitation models preserve the fine-scale detail that error-minimizing models blur \cite{yu2024diffcast, leinonen2023latent}, and it matters for flood risk, which depends on where the most intense rain falls.

\subsection{Joint structure and rare event tail}
\begin{figure}[!hbtp]
\centering
\figimg[0.7\textheight]{\linewidth}{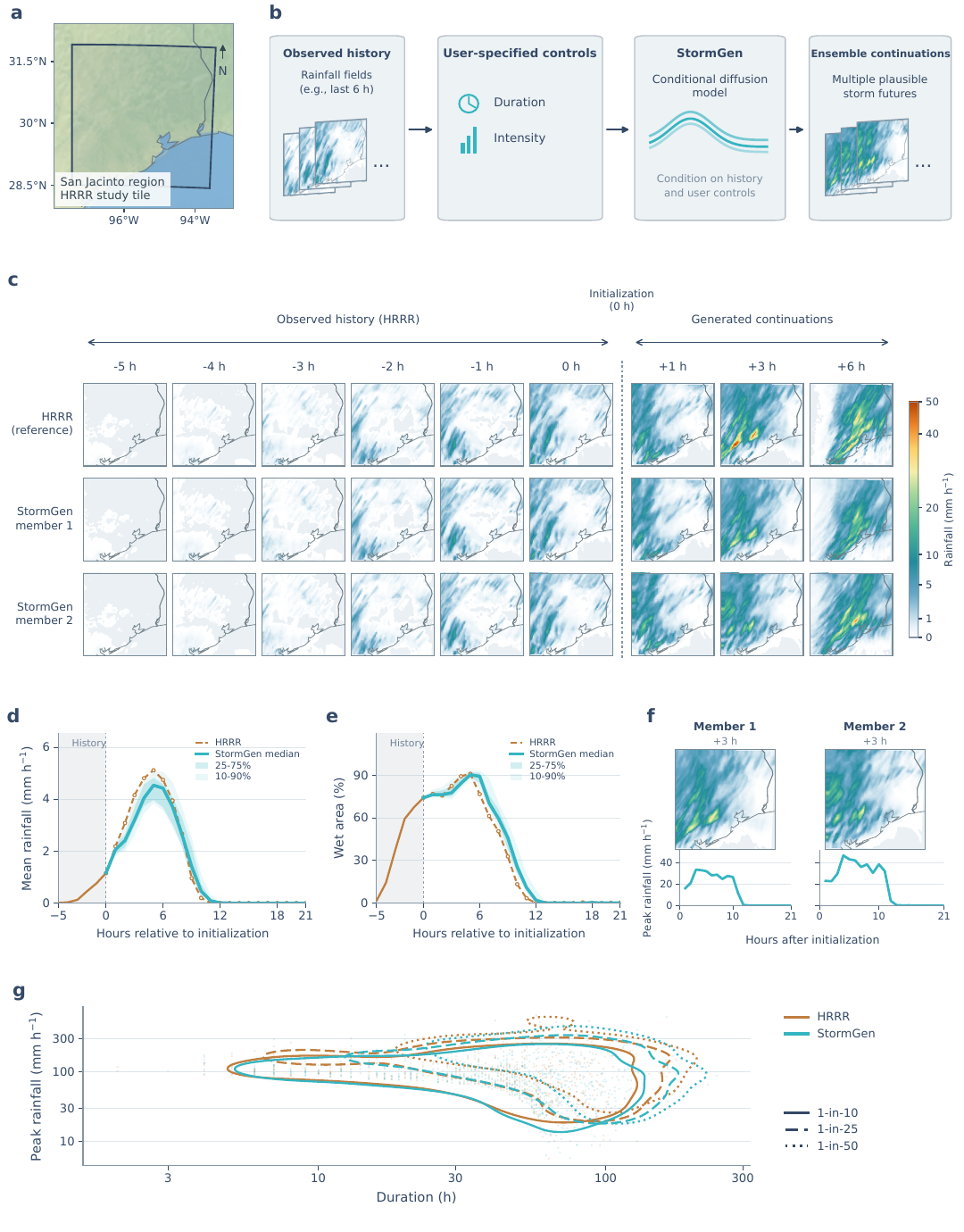}
\caption{\textbf{StormGen generates multiple plausible storm futures from an observed beginning.}
(\textbf{a}) Study tile over the San Jacinto region of southeast Texas.
(\textbf{b}) StormGen workflow. Six hours of observed rainfall (the seed) and user-specified controls (duration, intensity) condition the model, which returns an ensemble of complete continuations.
(\textbf{c}) Rainfall for a storm on 24 January 2023 in HRRR and two StormGen members. Columns left of the dashed line show the six-hour history ending at initialization (0h); columns to the right show continuations at +1, +3, and +6 hours.
(\textbf{d} and \textbf{e}) Domain-mean rainfall and wet-area percentage for HRRR (brown, dashed) and the median of 24 StormGen members (teal), with 25th to 75th and 10th to 90th percentile bands. Gray shading marks the history. Wet pixels have rainfall $\geq$ 0.05 mm\,h${-1}$.
(\textbf{f}) Controllability. Continuations of the same seed requested as a shorter (6-hour) and a longer (18-hour) event, with rainfall maps and time series.
(\textbf{g}) Joint distribution of duration and peak rainfall for HRRR and StormGen events, with density colors for the rarest 10, 4, and 2\% of storms (1-in-10, 1-in-25, 1-in-50). These classes are fractions of catalog events, not annual return periods.
Ensemble construction and event definitions are given in Materials and Methods.}
\label{fig:workflow}
\end{figure}

\begin{figure}[!hbtp]
\centering
\figimg[0.6\textheight]{\linewidth}{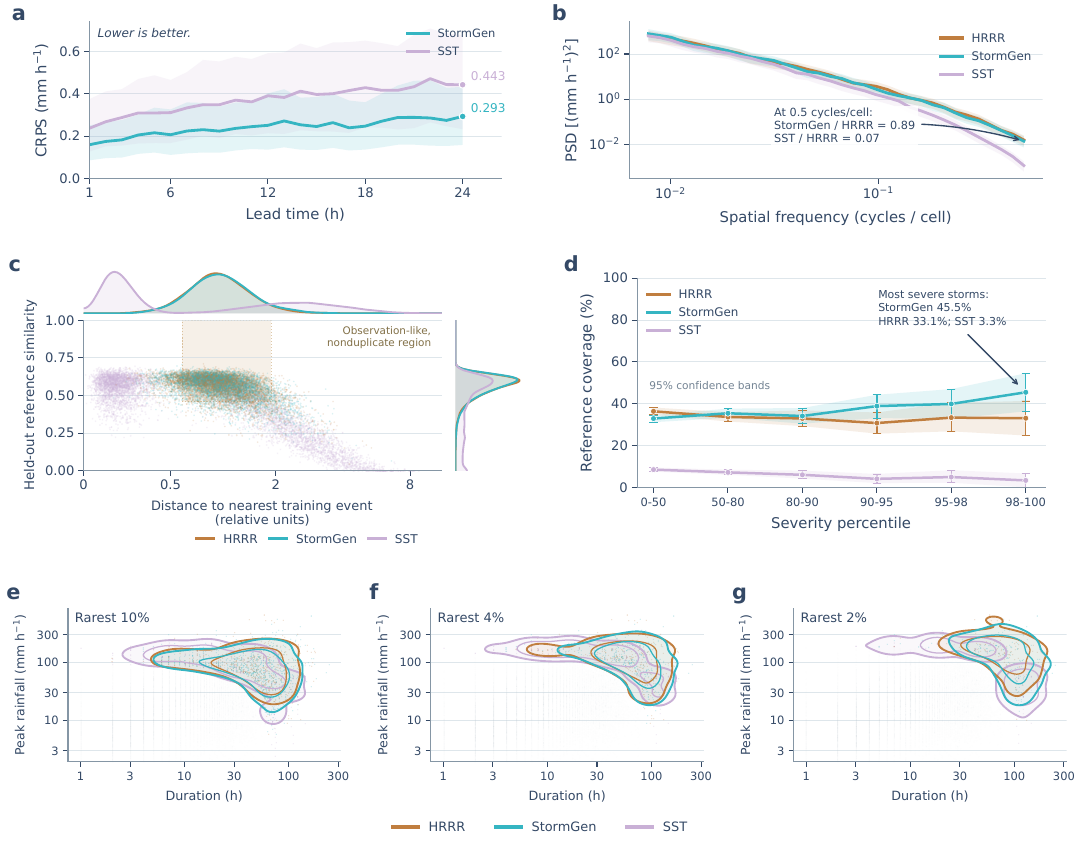}
\caption{\textbf{StormGen generates realistic, nonduplicate storms and improves coverage of the extreme tail.}
(\textbf{a}) Continuous ranked probability score (CRPS) of StormGen and SST ensemble against HRRR by lead time; lower is better. Lines show means and shading shows the spread across events.
(\textbf{b}) Spatial power spectra of rainfall fields. Higher spatial frequencies correspond to finer-scale features, such as localized area of heavy rain.
(\textbf{c}) Distance of each event to its nearest training event against its similarity to an independent held-out reference set, with marginal densities. The shaded region marks events that resemble observations without duplicating training events. These quantities measure duplication and resemblance, not the probability of physical realism.
(\textbf{d}) Percentage of reference events in each severity class that have a close counterpart in each catalog, at a fixed tolerance in storm-property space. Error bars show pointwise 95\% percentile bootstrap confidence intervals from 2,000 resamples of reference events within each severity class, with candidate catalogs and matching criteria held fixed.
(\textbf{e} and \textbf{g}) Joint distribution of duration and peak rainfall for the rarest 10, 5, and 2\% events of errors in each catalog, shown as density contours.
HRRR, StormGen, and SST are shown in brown, teal, and magenta, with 6,000 events each. Continuations that share a seed are dependent draws. Diagnostic definitions are given in Materials and Methods.}
\label{fig:fidelity}
\end{figure}

StormGen’s most consequential property is its behavior in the tail, where duration and intensity are both extreme and observations are scarcest. To examine it, we ranked the storms in each catalog by the joint rarity of their duration and peak intensity and isolated the rarest 10, 4, and 2\% (1-in-10, 1-in-25, and 1-in-50 storms). The rarest 2\% amounts to roughly 120 of 6,000 events: the long, intense storms that dominate flood risk, of which any single local records holds only a handful. Because flood risk depends on how duration and intensity properties combine, a catalog that reproduced them separately but paired them wrongly would misrepresent these storms. Within these classes, StormGen followed the observed joint distribution closely, extending it modestly toward longer events (Fig,~\ref{fig:workflow}g). Transposed storms also occupied the core of the observed tail, as expected of relocated real storms, but in the rarest classes they included a subsert of shorter, sharply peaked storms that is largely absent from the observations (Fig.~\ref{fig:fidelity}, e to g). Notably, StormGen achieved its agreement without copying its training data. Generated storms lay about as far from their nearest training examples as held-out observed storms do and resembled and independent reference set just as closely, whereas transposed storms were either near-duplicates of training storms or, where they departed from them, lost their resemblance to observed ones (Fig.~\ref{fig:fidelity}c).
This combination of novelty and realism allows StormGen to supply the extreme storms that a short record happens to miss. Taking the most severe 2\% of observed reference storms as a benchmark, we assessed how often each catalog contained a storm with similar characteristics. StormGen did so for about 45\% of them, compared with 33\% for an equally sized catalog of observed storms and 3\% for SST (Fig.~\ref{fig:fidelity}d). The confidence intervals of StormGen and the observed catalog overlap at every severity, so the generated catalog is best described as matching, rather than exceeding, a real catalog of the same size, which is itself notable for a model trained on 268 storms. Transposition, by contrast, covered few reference storms at any severity and fewer still toward the extremes, consistent with its dependence on the storms already in its regional catalog \cite{wright2020six}. StormGen’s modest lean toward longer, wetter storms nonetheless carries into an frequency estimate built from its catalog, a point we return to below. This suggests a role for generate catalogs in data-limited regions, where planning must account for severe storms beyond those captured in the available record. 

\section{Complete scenarios beyond the forecast horizon}
AI precipitation nowcasting models provide a useful comparison because they generate spatially detailed rainfall sequences, although their purpose is short-term prediction rather than record augmentation.
Models that resolve individual storms at the kilometer scale, such as DGMR \cite{ravuri2021dgmr} and NowcastNet \cite{zhang2023nowcastnet}, target horizons of only 90 minutes to 3 hours, whereas global models that run for days are too course for local flooding \cite{ravuri2021dgmr, zhang2023nowcastnet, lam2023learning, price2023gencast}. To test whether such a model could simply be extended to produce complete scenarios, we ran NowcastNet recursively to 24 hours over storms in a second region, the Southern Great Plains. Its rain bands drifted and faded after about six hours, whereas StormGen held each storm together from growth and decay (Fig.~\ref{figS:nowcast}), consistent with error compounding through that feedback loop, which StormGen avoids by generating a storm’s full lifecycle at once. Because StormGen also receives a storm’s requested duration and severity, the comparison therefore contrasts conditional scenario generation with recursively extended nowcasting, rather than providing a one-for-one test of forecast skill (Supplementary Text). The experiment also offers an early indication that StormGen’s approach carries over to a region beyond southeast Texas.

\section{Catchment-scale hazard for an unseen extreme: Hurricane Harvey}
\subsection{Storms beyond the training record}
Whether data-driven models can produce events more extreme than any in their training data remains an open question. A recent study found that an AI weather model trained without the strongest tropical cyclones could not forecast Category 5 storms, concluding that such models cannot extrapolate from weaker events \cite{sun2025grayswan}. A very consequential test for StormGen is therefore if StormGen can generate storms more extreme than those seen during training.A random train–test split does not directly test this ability, because both sets can contain similarly extreme storms. We therefore defined a boundary separating the most extreme combinations of storm duration and peak intensity from the remaining data and excluded all 96 storm windows beyond that boundary from training, including Hurricane Harvey (Fig~\ref{fig:beyond}a; Materials and Methods). StormGen then generated 32 members for each withheld storm, conditioned on its initial rainfall and on an extreme-storm regime defined from the training data, with no information about the storm’s own duration or intensity. 

Generated members populated the excluded region densely, extending beyond the boundary toward the longest and most intense withheld storms (Fig.~\ref{fig:beyond}a). For a typical withheld storm, 34.4\% of members fell within a factor of 1.25 of both its observed duration and its peak intensity, a fraction that varied from storm to storm but did not decline systematically with severity. Because the severity ranking and the exclusion rule were fixed before generation, this agreement cannot reflect tuning to the withheld storms.

\subsection{Hurricane Harvey from a single hour of rainfall}
\begin{figure}[!hbtp]
\centering
\figimg[0.7\textheight]{\linewidth}{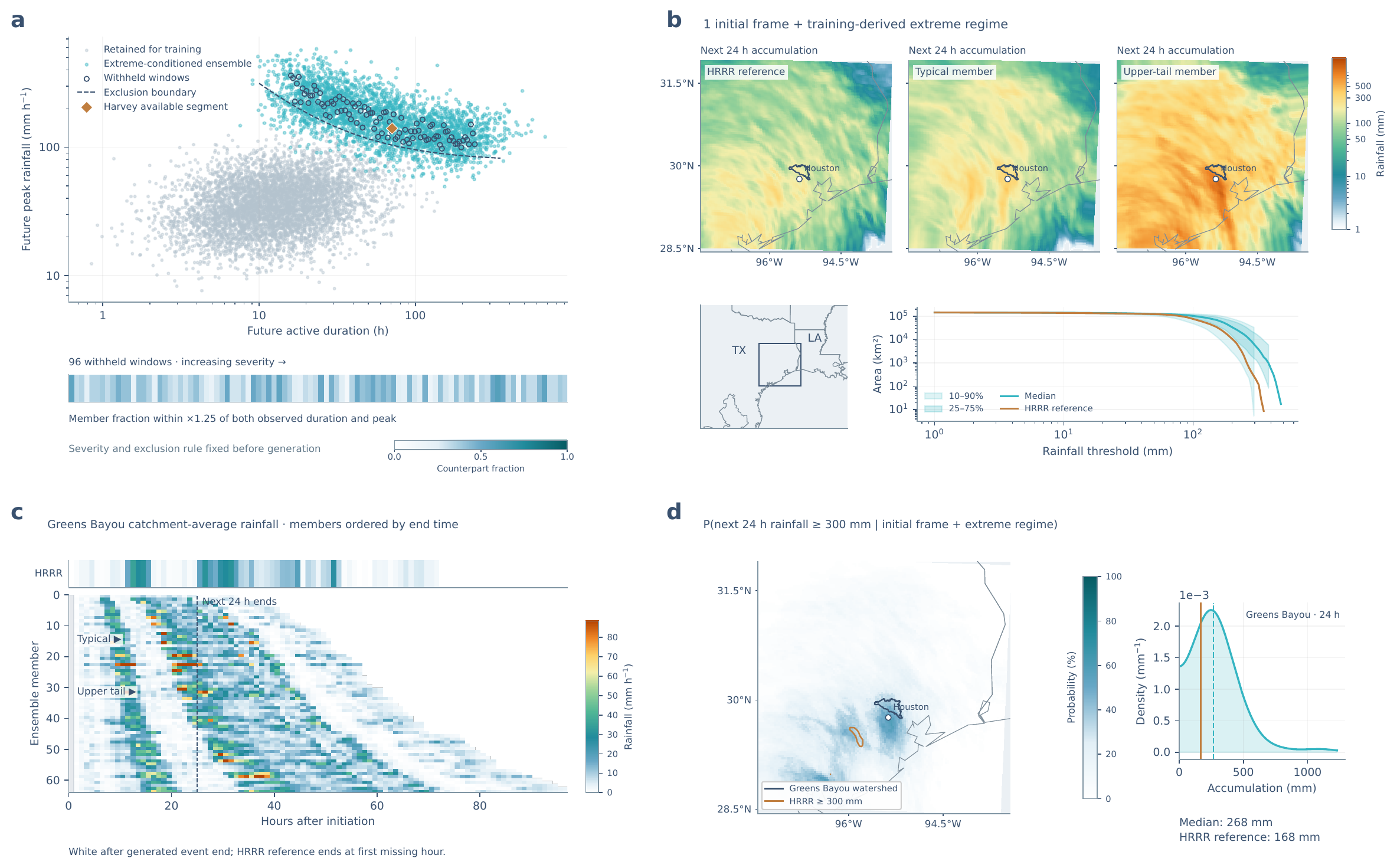}
\caption{\textbf{StormGen generates storms beyond its training record, including catchment-scale rainfall hazard for Hurricane Harvey.}
(\textbf{a}) Withheld-tail experiment. Storms beyond a fixed exclusion boundary in duration and peak rainfall (dashed line) were withheld before training. Gray points show retained storms, open circles the 96 withheld storms, teal points generated members, and the orange diamond the available Harvey segment. The strip below orders withheld storms by severity and shows, for each, the fraction of its 32 members within a factor of 1.25 of both its observed duration and peak.
(\textbf{b}) Next-24-hour accumulation for the HRRR reference, a typical member (nearest the ensemble median), and an upper-tail member (nearest the 90th percentile), initialized at 02:00 UTC on 26 August 2017 from a single hour of rainfall. Lower right, area receiving at least a given accumulation for HRRR (brown) and the ensemble median (teal), with 10th to 90th and 25th to 75th percentile bands.
(\textbf{c}) Hourly catchment-average rainfall over the Greens and Halls Bayou watersheds for HRRR (top) and 64 members ordered by duration. The dashed line marks the end of the 24-hour window; the HRRR record ends at its first missing hour.
(\textbf{d}) Probability that 24-hour rainfall exceeds 300 mm, with the watershed outline and the HRRR 300-mm contour ([ ]). Right, distribution of the 64 catchment-average 24-hour accumulations, with the ensemble median (dashed, 268 mm) and the HRRR reference (solid, 168 mm).
Catchment averages use area-weighted intersections computed in an equal-area projection. Details are given in Materials and Methods.}
\label{fig:beyond}
\end{figure}

Harvey is the most demanding case in the withheld set. We initialized StormGen at 02:00 UTC on 26 August 2017, near the time of landfall, from a single hour of HRRR rainfall (Fig.~\ref{fig:beyond}b). The 64 resulting continuations lasted between 32 and 96 hours, bracketing the 71 hours of reference record available, and, like Harvey, delivered their rain in repeated bursts rather than a single pulse (Fig.~\ref{fig:beyond}c). Over the following 24 hours, the members reproduced the broad pattern of the observed accumulation, and the area receiving a given depth tracked the reference up to about 100 mm, beyond which the ensemble placed more area at the highest totals (Fig. ~\ref{fig:beyond}b).
Summarized across members, those continuations yield the quantity on which flood management depends: the probability that rainfall over a specific watershed exceeds a critical depth (Fig.~\ref{fig:beyond}d). For the Greens and Halls Bayou watersheds north of downtown Houston (546 km$^2$), the ensemble placed the median 24-hour catchment-average rainfall at 268 mm, with a 1-in-10 chance of exceeding about 460 mm. The HRRR reference of 168 mm fell within this range but below its median, and members exceeded 300 mm over a broader areat than HRRR did, consistent with the lean toward wetter storms noted above. A conditional hazard estimate of this kind, obtained from one hour of observations for a storm unlike any in training, is the form of information that watershed-scale warnings and, aggregated across many storms, design standards and adaptation policy require. 

\section{Expanding rainfall scenarios for catchment-scale flood assessment}
\subsection{Rainfall timing and location shape downstream flood response}
\begin{figure}[!hbtp]
\centering
\figimg[0.7\textheight]{\linewidth}{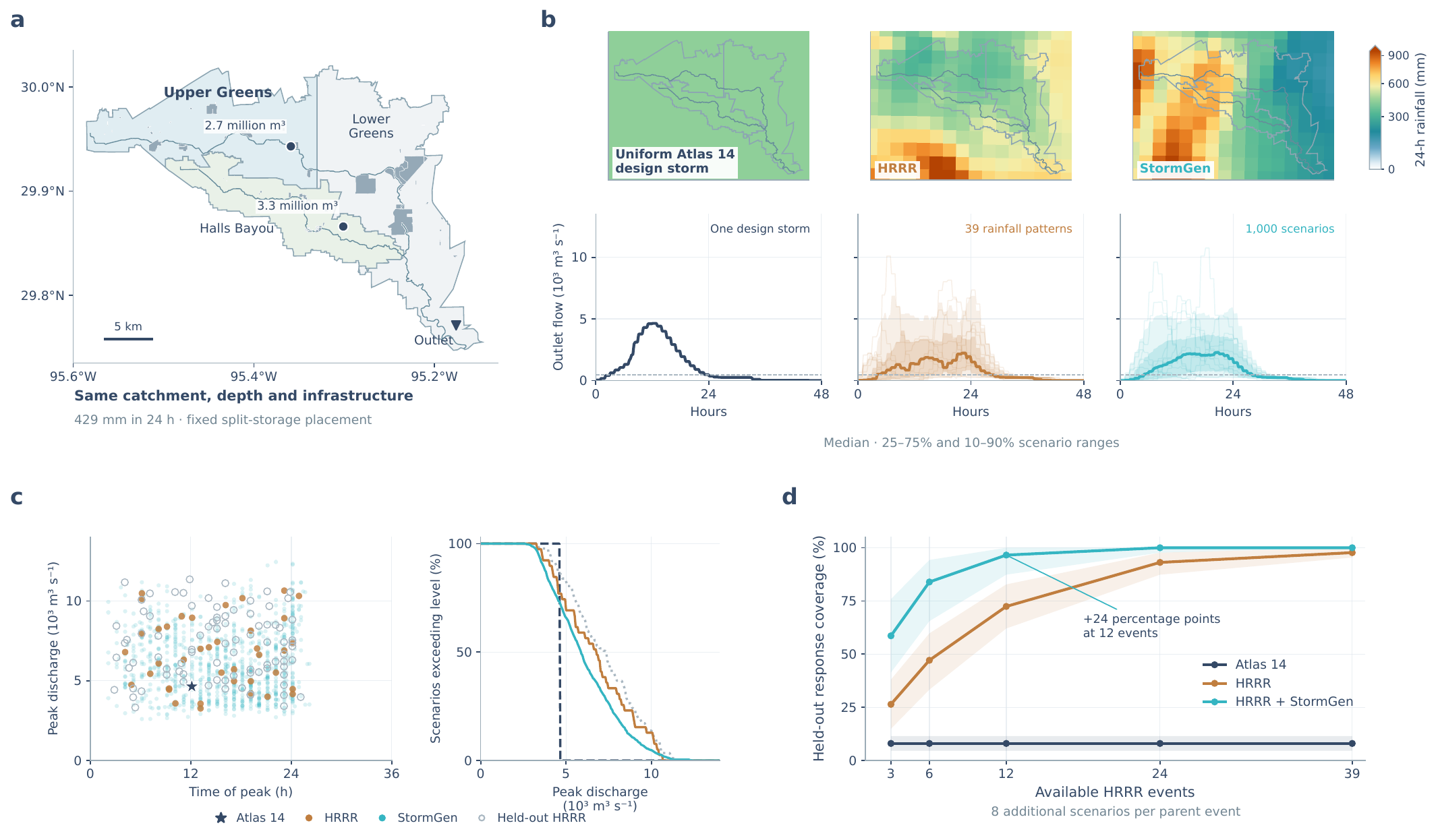}
\caption{\textbf{Rainfall-pattern variability and modeled flood responses at a common design depth.}
(\textbf{a}) Greens and Halls Bayou catchments with $6\times10^6$~m$^3$ of added storage divided between upper Greens (45\%) and Halls (55\%).
(\textbf{b}) Rainfall fields and hydrographs for a uniform Atlas~14 design storm, 39 HRRR patterns, and 1,000 additional scenarios. All have 429~mm catchment-average depth over 24~h, using the Atlas~14 100-year point estimate, under identical wet conditions and routing. Heavy curves show medians; bands show central 50\% and 80\% scenario ranges. The dashed line marks the assumed 450~m$^3$~s$^{-1}$ capacity.
(\textbf{c}) Peak-discharge magnitude and timing (left) and scenario exceedance fractions (right), compared with 87 held-out HRRR patterns (open circles and dotted curves).
(\textbf{d}) Held-out response coverage versus available HRRR events, with eight additional scenarios per parent. Counterparts match peak discharge within a factor of 1.25 and peak timing within 3~h. Curves show medians and bands the 10th--90th percentiles across 500 paired catalog-subsampling and held-out-bootstrap repetitions.}
\label{fig:rfa}
\end{figure}

The practical value of a storm generator depends on whether its additional scenarios reveal consequences that a limited historical record fails to represent. We examined this question in the combined Greens and Halls Bayou watersheds using a conceptual rainfall--runoff and detention-routing model. We compared three rainfall sources: a spatially uniform Atlas~14 design storm, 39 HRRR rainfall patterns from 2015--2017, and 312 StormGen scenarios conditioned on those historical events. All sources were evaluated with the same infrastructure: 6 million~m$^3$ of added storage, divided between upper Greens (45\%) and Halls (55\%) (Fig.~\ref{fig:rfa}a). Holding infrastructure fixed isolates the contribution of rainfall variability rather than differences in storage placement.

We standardized every rainfall scenario to a catchment-average depth of 429~mm over 24~h, using the Atlas~14 100-year, 24-hour point estimate at the catchment centroid. The uniform design storm applied one prescribed temporal distribution throughout the catchment; the HRRR and StormGen scenarios retained their spatial patterns and temporal sequences within the common evaluation window. StormGen received six hours of observed rainfall as conditioning history and generated the subsequent 24 hourly fields. The conditioning period was excluded from both the evaluated rainfall depth and the hydrologic simulation; the corresponding 24-hour HRRR continuation provided the historical comparison. Each scenario was multiplied by a single factor to achieve the common catchment-average depth, preserving its relative spatial and temporal structure. Wet antecedent conditions and routing parameters were identical across sources.  

The uniform design storm produced a modeled peak discharge of approximately 4,642~m$^3$~s$^{-1}$. Across the HRRR and StormGen expanded catalogs, the central 80\% of peak discharges spanned 4,090--10,135 and 3,655--9,029~m$^3$~s$^{-1}$, respectively, with corresponding variation in peak timing (Fig.~\ref{fig:rfa}b,c).
These differences arise because equal rainfall depths need not deliver runoff to the outlet at the same rate. We note that the catalog exceedance curves summarize the fraction of evaluated scenarios exceeding each peak-discharge level; they describe conditional scenario diversity. 

\subsection{Can generated storms fill gaps in a short historical catalog?}

We assessed downstream coverage using 87 HRRR rainfall patterns from 2018--2024, routed under the same rainfall-depth, infrastructure, and antecedent-condition assumptions. These evaluation patterns were excluded from the historical parent pool used to construct the additional scenarios. For historical subsets containing 3, 6, 12, 24, or 39 events, we compared the original catalog with the same catalog augmented by eight additional scenarios per parent event. In the current illustration, these additional scenarios are HRRR-derived perturbations representing the proposed StormGen augmentation experiment. The Atlas~14 baseline remained one fixed design storm. A held-out response was considered covered when a catalog contained at least one counterpart with peak discharge within a factor of 1.25 and peak timing within three hours. We repeated catalog subsampling and held-out-event bootstrapping 500 times, reporting median coverage and its 10th--90th percentile range (Fig.~\ref{fig:rfa}d).

With 12 historical events, median coverage was 72.4\% for HRRR alone and 96.6\% after augmentation, a difference of 24.1 percentage points between the catalog medians. The corresponding 10th--90th percentile ranges were 62.1--82.8\% and 87.4--100\%, respectively. The uniform design storm covered a median of 8.0\% of held-out responses. As the historical catalog increased to 39 events, coverage reached 97.7\% for HRRR alone and 100\% after augmentation. At this catalog size, the paired coverage gain had a median of 2.3 percentage points and a 10th--90th percentile range of 0--4.6 percentage points. The additional scenarios therefore supplied counterparts to combinations of peak discharge and peak timing missing from smaller historical catalogs, while the incremental gain narrowed as the historical catalog grew from 12 to 39 events.

The expanded catalog increased coverage of held-out flood responses, with larger gains when historical support was limited and smaller gains as the historical catalog grew. Because storage placement was held fixed, this experiment broadened the assessment of flood response without testing whether augmentation improves an engineering decision. This distinction is central to evaluating StormGen’s contribution to adaptation: representing more possible flood responses does not by itself establish better protection. 

\section{Discussion}
StormGen shows how a generative model can expand the range of storm scenarios available from a short rainfall record. Across the evaluated distribution, it reproduced key features of storm duration, intensity and spatial organization while generating events distinguishable from its training examples. More importantly, when the most severe storms were excluded from training, StormGen generated continuations extending into the withheld duration–intensity regime, including Harvey-scale events conditioned on initial rainfall and requested storm characteristics. These results suggest that the model learned relationships that support some extrapolation beyond the retained catalog. The distinction from stochastic storm transposition is specific: transposition can produce previously unobserved local extremes by relocating storms from a wider regional archive, whereas StormGen can also generate variations in their evolution. Its greater coverage of severe reference events relative to SST therefore supports a complementary role for learned generation in exploring sparsely sampled combinations of storm properties. 

The practical importance of this capability lies in the consequences of those additional scenarios. StormGen preserves rainfall’s spatial and temporal evolution, allowing generated events to drive watershed and infrastructure models directly. In the catchment experiment, storms constrained to the same 24-hour basin-average rainfall depth produced more than a twofold range in modeled peak discharge. Within the assumptions of the hydrologic model, this demonstrates how event totals can conceal consequential differences in rainfall placement and timing.
Two limitations identify clear priorities for further development. First, StormGen tends to generate storms that are longer and wetter than those in the reference catalog. This highlights the challenge of expanding the range of plausible extremes while retaining the observed balance of storm duration and rainfall amount. The additional scenarios can support conditional stress testing, but their frequency and magnitude require calibration before they are used to estimate return levels or exceedance probabilities. An important next step is therefore to reduce these biases without suppressing the diversity that makes generation useful.

Second, StormGen requires historical rainfall sequences that capture complete storm lifecycles for training. These sequences provide the information needed to learn how rainfall growth, movement and decay relate to one another, but their availability limits where the framework can currently be developed and evaluated. StormGen can expand the scenarios represented in an existing record; it cannot eliminate the need for a sufficiently informative training archive. This requirement concerns model training, whereas subsequent generation can begin from an initial rainfall history and requested storm characteristics. Broader training across regions and rainfall products offers a potential route toward use in places with shorter records, although the reliability of that transfer remains to be established.

\FloatBarrier
% ---------------------------------------------------------------------
% References
% ---------------------------------------------------------------------
\clearpage
\IfFileExists{sciencemag.bst}%
  {\bibliographystyle{sciencemag}}%
  {\bibliographystyle{unsrt}}
\bibliography{stormgen_refs}

% ---------------------------------------------------------------------
% Back matter (required sections for Science)
% ---------------------------------------------------------------------
\subsection*{Supplementary Materials}
Materials and Methods \\
Figs. S1 to S5

% =====================================================================
% Supplementary Materials (Science places Materials and Methods here)
% =====================================================================
\clearpage
\setcounter{figure}{0}
\setcounter{table}{0}
\setcounter{equation}{0}
\renewcommand{\thefigure}{S\arabic{figure}}
\renewcommand{\thetable}{S\arabic{table}}
\renewcommand{\theequation}{S\arabic{equation}}

\begin{center}
{\Large Supplementary Materials for}\\[6pt]
{\large\bfseries Generated storm lifecycles fill the rare-event tail of short rainfall records}\\[6pt]
Valerie Tsao, Nathaniel Chaney, Manolis Veveakis, Marta Zaniolo
\end{center}

\section*{Materials and Methods}
\subsection*{S1. Scope and data sources}

StormGen generates spatial rainfall sequences conditioned on an initial rainfall history and supplied storm descriptors. The development workflow uses six hourly history fields and represents the continuation jointly in space and time.

The main rainfall archive is a regional High-Resolution Rapid Refresh (HRRR) tile covering southeast Texas and the San Jacinto region. HRRR provides numerical weather prediction fields rather than direct rainfall observations. Catchment calculations for the Hurricane Harvey reference case use the combined Greens Bayou and Halls Bayou watersheds within this larger tile.

Separate analyses use Stage IV radar--gauge precipitation and RainyDay catalogs to examine agreement with NBM precipitation probabilities and rainfall-frequency estimates. These analyses use different basins, data partitions, and sampling units from the HRRR development workflow and are described separately below.

\subsection*{S2. HRRR rainfall archive and spatial representation}

\paragraph{Rainfall archive.}
The archive contains hourly HRRR surface total precipitation at forecast lead hour one, accumulated over forecast hours 0--1. Accumulation starts range from 1 January 2015 at 00:00 UTC through 31 December 2024 at 23:00 UTC.

Each field represents a one-hour water-equivalent depth. Its numerical value in millimeters is therefore equal to its hourly mean rate in millimeters per hour; it is not an instantaneous rainfall measurement. Unless stated otherwise, timestamps refer to accumulation interval ends. HRRR has a nominal 3-km grid.

\paragraph{Regional tile and catchment weights.}
The native projected crop contains $128\times128$ cells at nominal 3-km spacing, covering approximately $384\times384$ km. Its geographic envelope is approximately $28.42419$--$31.91518^{\circ}$N and $97.45987$--$93.41859^{\circ}$W. The tile includes the San Jacinto region but extends well beyond the San Jacinto watershed.

Regional means use the full tile with a uniform projected cell-area approximation of $9\,\mathrm{km^2}$. Catchment means instead use the area of each grid cell intersecting the catchment polygon. These intersections are calculated in an equal-area projection.

\subsection*{S3. Regional episode identification}

Regional rainfall episodes are identified using hysteresis in tile-mean rainfall and wet-area fraction. An episode begins when the mean rate reaches $0.02\,\mathrm{mm\,h^{-1}}$ or the wet fraction reaches 0.002. It remains active while the mean rate is at least $0.005\,\mathrm{mm\,h^{-1}}$ or the wet fraction is at least 0.0005.

An episode may contain up to six consecutive hours below both continuation thresholds. Retained episodes require a confirmed seven-hour inactive interval before onset and after the last active hour. They must contain at least eight hours, providing six history hours and at least two continuation hours. The prepared catalog contains 389 episodes.

The wet-pixel cutoff is calibrated from positive rainfall in the permitted training interval and is approximately $0.001\,\mathrm{mm\,h^{-1}}$. This detection threshold is distinct from the $0.05\,\mathrm{mm\,h^{-1}}$ footprint threshold used for the January 2023 reference example.

These episodes describe regional rainfall spells. An episode may contain several rain bands or convective systems and should not be interpreted as a single tracked convective cell.

\subsection*{S4. Conditional diffusion model}

\paragraph{Learning target.}
Let $R_t(j)$ denote hourly mean rainfall rate in cell $j$, and let
\[
\mathcal H=(R_{-5},\ldots,R_0)
\]
denote the six-hour history ending at initialization. StormGen represents
\begin{equation}
p_{\theta}\!\left(R_{1:L}\mid\mathcal H,c\right),
\end{equation}
where $L$ is the requested continuation length and $c$ contains the available storm descriptors.

The future is represented jointly in space and time. Each ensemble member starts from an independently seeded noise realization, while the history remains fixed across members. The denoiser predicts the clean continuation.

\paragraph{Architecture and conditioning.}
The development implementation uses $8\times8$ spatial patches, six transformer layers, an embedding width of 128, and four attention heads. Spatial attention operates within $4\times4$-patch windows, and temporal attention uses windows of 32 positions. A local decoder contains three $3\times3$ convolutional layers with hidden width eight.

The model conditions on rainfall history, spatial and temporal coordinates, cadence, and numerical storm descriptors. The descriptors are elapsed span, active duration, peak rate, wet-pixel mean rate, mean depth, and maximum wet fraction. They are transformed by $\log(1+c)$ and standardized using training statistics. Explicit availability masks distinguish supplied descriptors from unavailable ones.

The diffusion and transformer formulations follow the foundations described by \cite{ho2020denoising} and \cite{vaswani2017attention}.

\paragraph{Rainfall representation and diffusion loss.}
Rainfall is encoded using
\begin{equation}
\mathcal T(r)=
\begin{cases}
-1, & r=0,\\[2pt]
\sqrt{r/r_0}, & r>0,
\end{cases}
\qquad
r_0=1\,\mathrm{mm\,h^{-1}}.
\end{equation}
The decoded rate is
\begin{equation}
r=r_0\max(z,0)^2.
\end{equation}

For the represented clean future $z_0=\mathcal T(R_{1:L})$, forward corruption is
\begin{equation}
z_{\tau}=\alpha_{\tau}z_0+\sigma_{\tau}\epsilon,
\qquad
\epsilon\sim\mathcal N(0,I),
\end{equation}
with
\begin{equation}
\alpha_{\tau}=\cos(\pi\tau/2),
\qquad
\sigma_{\tau}=\sin(\pi\tau/2).
\end{equation}
Sampled diffusion times are clipped to $[0.001,0.999]$.

The clean-prediction loss is
\begin{equation}
\mathcal L_{\mathrm{denoise}}
=
\mathbb E\left[
w(\tau)
\left\|
f_{\theta}(z_{\tau},\tau,\mathcal H,c)-z_0
\right\|^2_{\mathrm{valid}}
\right],
\end{equation}
where
\begin{equation}
w(\tau)=\min(\sigma_{\tau}^{-2},5).
\end{equation}
The loss is evaluated over valid space--time entries.

The spatial reconstruction term has coefficient 0.1. Control-consistency and endpoint penalties have coefficients 0.25 and 1, respectively, ramped over the first 1,000 updates. Every fourth update uses a free-generation trajectory with 64 model evaluations, with gradients propagated through the final four evaluations.

\paragraph{Training and sampling.}
The development run used a fresh AdamW optimizer for 12,000 updates. Each update accumulated four one-parent microbatches. The learning rate increased to $5\times10^{-5}$ over the first 500 updates, then followed a cosine decay to $2.5\times10^{-6}$. Weight decay was 0.01, the gradient-norm clipping threshold was 1, and the exponential-moving-average decay was 0.995.

Training used FP32 arithmetic, random seed 20260915, six history hours, and proposed continuation lengths of 24, 36, or 72 hours. Sampled validation was performed every 2,000 updates with two members. The selected checkpoint was update 12,000.

Final development evaluation used four members and 64 joint clean-prediction DDIM evaluations, with guidance scale 1. Checkpoint selection considered sampled descriptor, endpoint, and spatial-artifact diagnostics, together with CRPS when storm controls were unavailable, rather than relying on denoising loss alone.

\subsection*{S5. Rainfall summaries and ensemble scoring}

\paragraph{Rainfall summaries.}
Let $a_j$ denote either grid-cell area or the area intersecting a catchment, and define
\[
w_j=\frac{a_j}{\sum_k a_k}.
\]
For a verification window $\mathcal W$ and hourly cadence $\Delta t=1\,\mathrm h$, the area-mean rainfall rate, wet fraction, mean depth, and peak rate are
\begin{align}
\bar R_t
&=\sum_j w_jR_t(j),\\
f_t(u)
&=\sum_jw_j\mathbf 1\{R_t(j)\geq u\},\\
B
&=\Delta t\sum_{t\in\mathcal W}\bar R_t,\\
P
&=\max_{t\in\mathcal W,j}R_t(j).
\end{align}
Here $u$ is the wet-pixel threshold for the analysis. Mean depth $B$ is expressed in millimeters, while $P$ is expressed in millimeters per hour.

When the denominator is positive, the event wet-pixel mean is
\begin{equation}
W=
\frac{
\sum_{t\in\mathcal W,j}
w_jR_t(j)\mathbf 1\{R_t(j)\geq u\}
}{
\sum_{t\in\mathcal W,j}
w_j\mathbf 1\{R_t(j)\geq u\}
}.
\end{equation}

Elapsed event span includes intervening quiet hours and is distinct from active duration. Continuation summaries exclude the supplied history; complete-event summaries include it. A sequence that remains active at its final available frame is censored rather than treated as having ended.

\paragraph{Continuous ranked probability score.}
For ensemble values $x_1,\ldots,x_M$ and a reference value $y$, the empirical ensemble CRPS is
\begin{equation}
\mathrm{CRPS}
=
\frac{1}{M}\sum_{m=1}^{M}|x_m-y|
-
\frac{1}{2M^2}
\sum_{m=1}^{M}\sum_{n=1}^{M}|x_m-x_n|.
\end{equation}
CRPS is calculated on the common valid grid at each lead, spatially averaged within a case, and then averaged with equal parent-event weights. Lower values indicate smaller probabilistic error.

This expression is the empirical-ensemble score and does not include a finite-ensemble bias correction. Comparisons therefore require a common ensemble size. Scores calculated with controls derived from a reference continuation describe conditional scenario error rather than forecast error using history alone. The scoring framework is described by \cite{gneiting01032007}.

\subsection*{S6. HRRR reference cases}

\paragraph{January 2023 sequence.}
The January 2023 reference sequence is initialized at 15:00 UTC on 24 January 2023. Its six history fields end at 10:00 through 15:00 UTC, followed by 21 continuation hours ending at 12:00 UTC on 25 January.

Hourly summaries use domain-mean rainfall and wet-area percentage. The wet-area calculation uses $u=0.05\,\mathrm{mm\,h^{-1}}$. Spatial comparisons use the same geographic extent and rainfall color limits throughout the sequence.

\paragraph{Hurricane Harvey reference accumulation.}
The Harvey reference calculation uses the hourly HRRR field ending at 02:00 UTC on 26 August 2017 as its temporal origin. The next-24-hour accumulation includes intervals ending at 03:00 UTC on 26 August through 02:00 UTC on 27 August.

For each grid cell,
\begin{equation}
A_{24}(j)
=
\Delta t\sum_{t=1}^{24}R_t(j).
\end{equation}
The combined Greens--Halls catchment has an area of $546.28\,\mathrm{km^2}$. Its basin-average accumulation is
\begin{equation}
B_{24}=\sum_jw_jA_{24}(j),
\end{equation}
where the weights use fractional cell intersections with the combined catchment polygon.

Over this 24-hour window, basin-average depth is 168.03 mm and maximum grid-cell depth is 353.10 mm. Across the full regional tile, $254.51\,\mathrm{km^2}$ receives at least 300 mm. Basin-average depth over the available 71-hour reference window is 593.01 mm.

For a depth threshold $q$, the regional area receiving at least that depth is
\begin{equation}
E(q)=\sum_j a_j\mathbf 1\{A_{24}(j)\geq q\},
\end{equation}
with $a_j$ expressed in square kilometers. This regional area is calculated over the full native tile, separately from the Greens--Halls basin-average depth.

\subsection*{S7. NBM and RainyDay catalog-probability agreement}

\paragraph{Reference precipitation and basins.}
This analysis uses native Stage IV radar--gauge precipitation. The basins are East Fork San Jacinto above Cleveland, USGS 08070000, and Caney Creek near Splendora, USGS 08070500.

Their polygon areas are approximately 844.40 and 265.16 $\mathrm{km^2}$, respectively. The spatial analysis uses 85 native Stage IV cells with positive East Fork intersection and 37 with positive Caney Creek intersection.

The completed comparison includes two catalog arms: observed-only and observed-plus-RainyDay. It contains 93 basin cases, comprising 46 Caney Creek cases and 47 East Fork cases, grouped into 47 cross-basin meteorological clusters.

\paragraph{NBM product and temporal alignment.}
The operational comparator is the archived CONUS NBM Quantile Mapping and Dressing probabilistic quantitative precipitation product. The analysis uses direct native 24-hour exceedance messages, with probabilities converted from percent to $[0,1]$.

Thresholds common to the inspected archive are 0.254, 6.35, 12.7, 25.4, 50.8, 76.2, and 101.6 mm. The principal agreement thresholds are 25.4 and 50.8 mm.

The comparison is made six hours after physical event onset, $T+6\,\mathrm h$, with both estimates targeting the fixed interval $[T,T+24\,\mathrm h]$. Eligible onsets align with 00, 06, 12, or 18 UTC native boundaries.

Forecast eligibility is determined by archived availability at or before the decision cutoff, rather than by nominal model cycle alone. The archive object's availability timestamp serves as a reproducible proxy for accessible information. The selected cycle, target interval, and availability margin are retained for each case.

The common cutoff does not make the conditioning information identical. NBM incorporates atmospheric-model information, while catalog retrieval explicitly uses the observed six-hour rainfall prefix. The analysis measures agreement between estimates based on these different information sources.

\paragraph{Data partitions.}
The Stage IV training interval begins on 1 January 2002 and ends on 14 March 2021 at 19:00 UTC. Validation extends from that boundary to 8 August 2023 at 10:00 UTC, and testing extends to 1 January 2026. These partitions apply to the NBM study and are separate from the HRRR development archive.

\paragraph{Conditional catalog probabilities.}
The two arms share the same permitted observed catalog. RainyDay augmentation uses training-source storms and stochastic storm transposition, which relocates rainfall fields while retaining their within-event structure \cite{wright2020six}.

Candidate histories are compared with the query's first six hours using normalized hyetographs and a fixed dynamic-time-warping distance. A magnitude-ratio screen of 0.3--3.0 is applied, and each arm selects $K=50$ analogs subject to source-group and diversity constraints. The source-group cap, warping window, and diversity thresholds are fixed in the analysis configuration.

For each selected analog, the query's known first-six-hour depth is added to the analog's continuous rainfall over hours 6--24. This continuation includes any subsequent event within the target window; it is not truncated at the end of the initially detected episode. The catalog probability is the fraction of selected totals strictly exceeding the native NBM threshold.

Rainfall depth is remapped before thresholding. Conservative polygon overlap is the primary mapping, and nearest-cell mapping is used for sensitivity analysis. Missing temporal or geographic support is not replaced with zero rainfall.

\paragraph{Agreement statistic.}
Let $p_{ijq}^{\mathrm{NBM}}$ denote the NBM probability for case $i$, cell $j$, and threshold $q$, and let $p_{aijq}^{\mathrm{cat}}$ denote the corresponding probability from catalog arm $a$. The case-level discrepancy is
\begin{equation}
D_{aiq}
=
\sum_jw_{ij}
\left(
p_{ijq}^{\mathrm{NBM}}-p_{aijq}^{\mathrm{cat}}
\right)^2.
\end{equation}
The differences are squared before spatial averaging, so discrepancies of opposite sign cannot cancel.

The augmentation effect is
\begin{equation}
\Delta_{iq}
=
D_{\mathrm{RainyDay},iq}
-
D_{\mathrm{observed},iq}.
\end{equation}
A negative value indicates closer agreement with NBM. It does not establish greater accuracy against realized rainfall or identify either estimate as the true conditional distribution.

Area-averaged cell probabilities describe the expected fraction of basin area exceeding a local rainfall threshold. They are not the probability that basin-average rainfall exceeds that threshold.

\paragraph{Uncertainty and dependence.}
The comparison uses 2,000 paired bootstrap resamples of complete cross-basin, 48-hour storm clusters. Both catalog arms use the same resampled clusters, and pooled point estimates weight the two basins equally. Cells and generated descendants are not treated as independent storms.

An additional 500 selected-source-group resamples characterize finite-catalog uncertainty conditional on the selected neighborhood. These intervals do not include uncertainty from refitting a generative model.

\paragraph{Mapping sensitivity.}
The mapping analysis compares augmentation effects under conservative and nearest-cell depth mappings. A main-threshold effect is treated as practically null when both mean effects lie within $\pm0.0025$.

Otherwise, directional stability requires the two effects to have the same sign and their difference to be smaller than the larger of the conservative effect magnitude and 0.0025. The nonsensitivity rule must be satisfied in at least 90\% of cluster-bootstrap draws. Probability profiles are also checked for monotonicity, duplicate messages, and consistent collocation across thresholds.

The completed observed-only and observed-plus-RainyDay comparison is shown in Figs.~\ref{figS:nbm_agreement} and~\ref{figS:nbm_mapping}.

\subsection*{S8. Long-record RainyDay rainfall-frequency benchmark}

The RainyDay frequency benchmark uses East Fork Stage IV rainfall from water years 2003--2020. Its source catalog contains 366 storms. The simulation comprises 1,000 years per realization and 100 realizations.

The candidate occurrence rate is
\begin{equation}
\lambda_{\mathrm{candidate}}
=
\frac{366}{18}
\simeq20.33
\quad\text{storms per year}.
\end{equation}
A separate common tail procedure retains three exceedances per year. This tail-exceedance rate is distinct from the candidate occurrence rate.

Basin-average rainfall is calculated from fractional cell intersections before extracting maximum rolling 24-hour depths. Averaging separately maximized grid-cell depths would give a different quantity. The benchmark analyzes rolling 24-hour maxima and 72-hour depths under the same tail-estimation convention.

The reference frequency curve is fitted to water years 2003--2020, the same source period used for the RainyDay catalog. The comparison therefore describes frequency estimates conditional on this archive, rather than an evaluation against an independent held-out climate record.

Variation among the 100 realizations and water-year bootstrap intervals are reported separately. Realization spread reflects variability among the simulations, while the water-year bootstrap assesses sensitivity to the sampled reference years. The simulated years do not extend the 18 years of observed climate exposure.
The benchmark is shown in Fig.~\ref{figS:rainyday_frequency}.

\section*{Supplementary figures}
\renewcommand{\thefigure}{S\arabic{figure}}
\setcounter{figure}{0}

\begin{figure}[!ht]
\centering
\includegraphics[width=\linewidth]{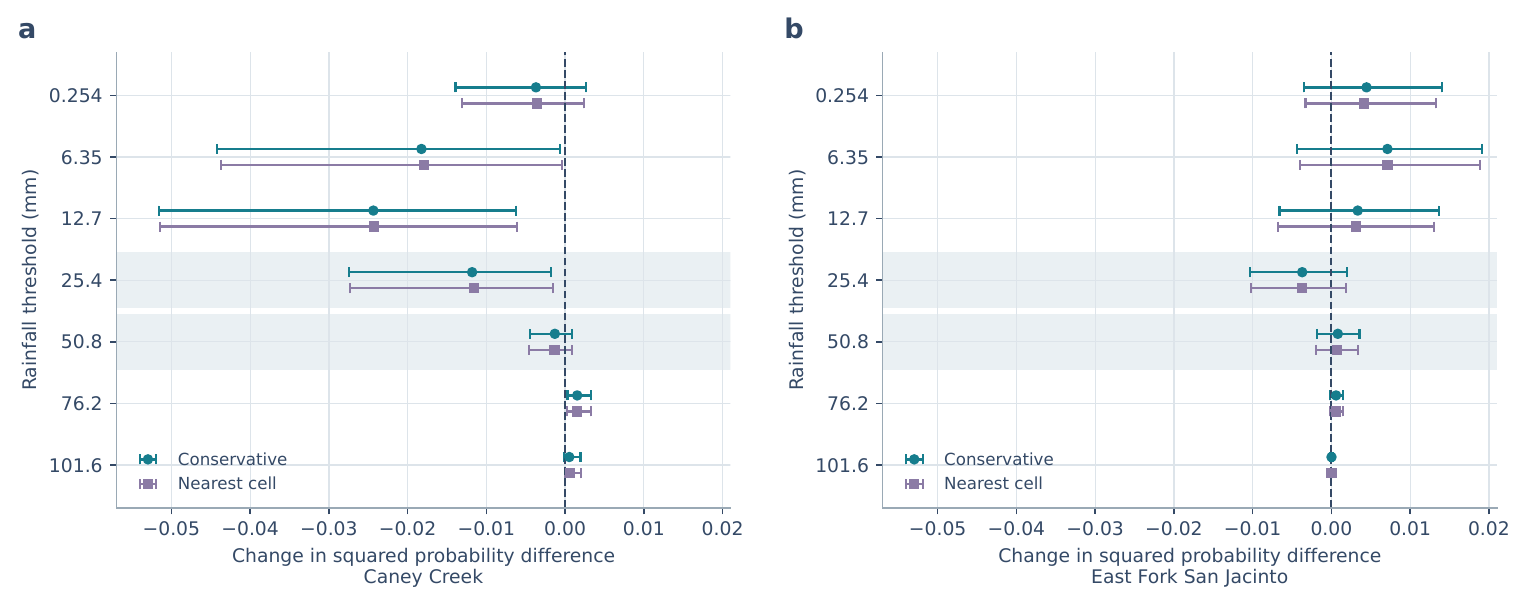}
\caption{\textbf{Agreement between conditional catalog probabilities and NBM.}
Panels show the change in agreement after RainyDay augmentation for
(\textbf{a}) Caney Creek and (\textbf{b}) East Fork San Jacinto,
using conservative and nearest-cell remapping.
The plotted quantity is
$D_{\mathrm{observed+RainyDay}}-D_{\mathrm{observed}}$, where $D$ is the basin-area-weighted mean squared difference between catalog and NBM cellwise exceedance probabilities.
Negative values indicate closer agreement with NBM.Points show mean changes, and whiskers show pointwise 95\% intervals from 2,000 paired bootstrap resamples of meteorological storm clusters.
Shading marks the prespecified primary thresholds of 25.4 and 50.8 mm; the remaining thresholds provide diagnostic comparisons.
The evaluation includes 93 basin cases---46 in Caney Creek and 47 in East Fork---from 47 cross-basin storm clusters. Probabilities are compared six hours after event onset and refer
to rainfall over the same fixed onset-to-24-hour interval. This comparison measures agreement between probability estimates, not verification against observed rainfall.}
\label{figS:nbm_agreement}
\end{figure}
\clearpage

\begin{figure}[!ht]
\centering
\includegraphics[width=\linewidth]{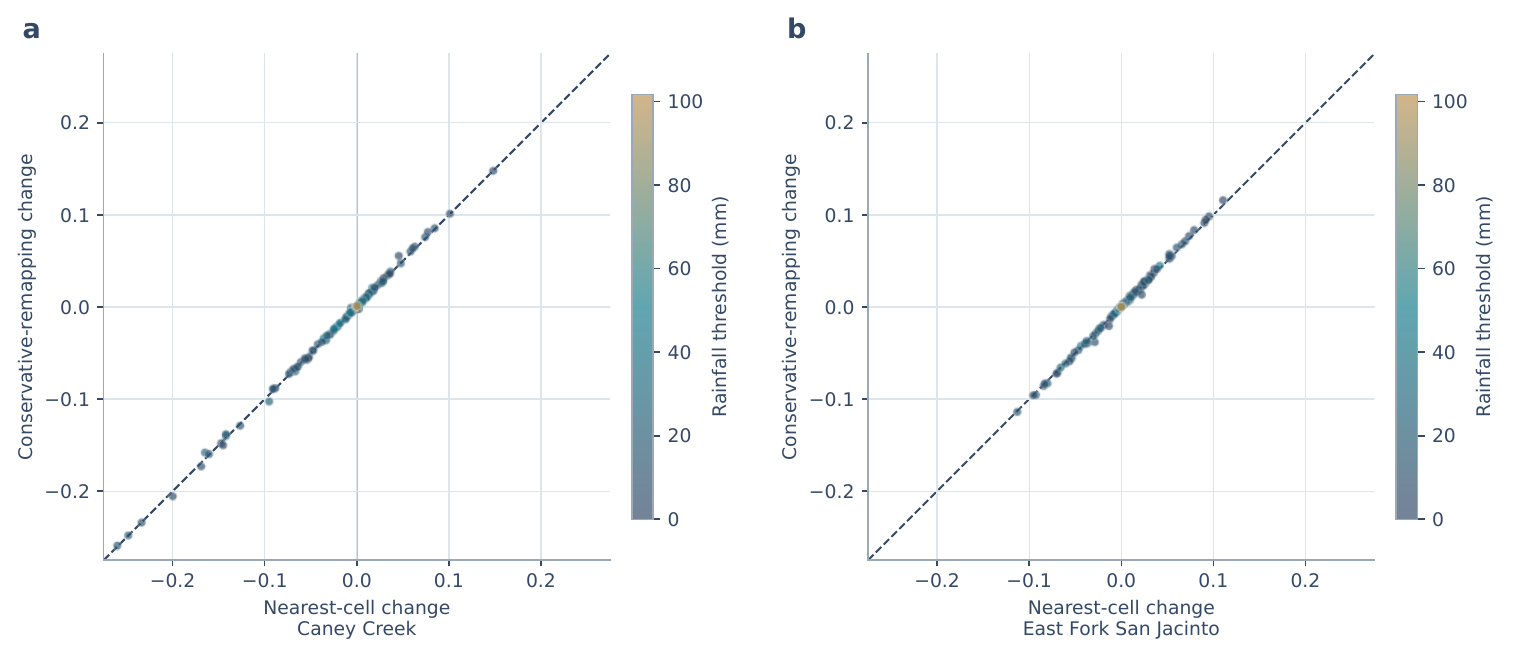}
\caption{\textbf{Sensitivity of catalog--NBM agreement to spatial remapping.}
Each point compares the effect of RainyDay augmentation for the same
basin case and rainfall threshold under nearest-cell and conservative
remapping. Panels show (\textbf{a}) Caney Creek and
(\textbf{b}) East Fork San Jacinto. Both coordinates represent $D_{\mathrm{observed+RainyDay}}-D_{\mathrm{observed}}$, with negative values indicating closer agreement with NBM,
as in Fig.~\ref{figS:nbm_agreement}. Color identifies the rainfall threshold, and the dashed diagonal marks identical augmentation effects under the two mappings.
All seven available thresholds are included. Rainfall depths are remapped before calculating exceedances. Multiple thresholds and basin cases can belong to the same meteorological storm, so the plotted points are not independent replicates.}
\label{figS:nbm_mapping}
\end{figure}
\clearpage

\begin{figure}[!ht]
\centering
\includegraphics[width=\linewidth]{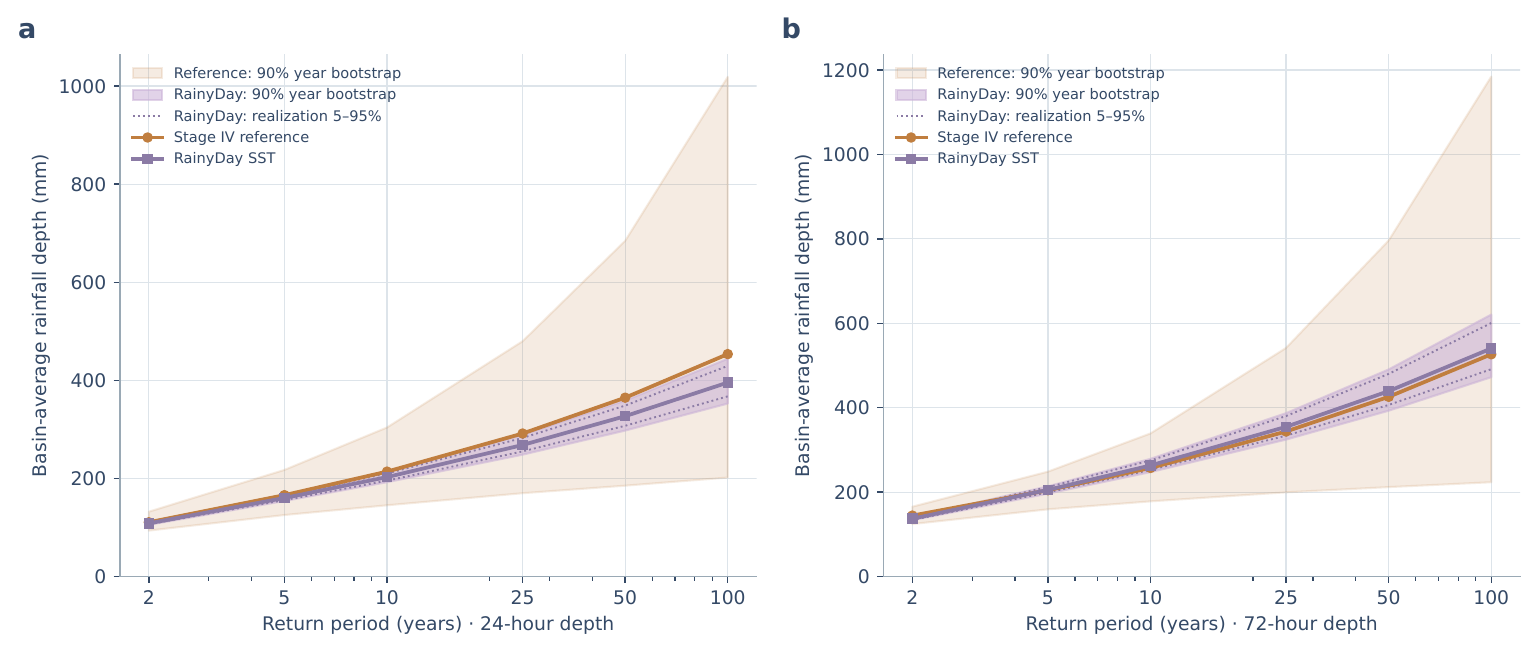}
\caption{\textbf{Stage IV and RainyDay rainfall-frequency estimates under a common tail model.}
Panels show (\textbf{a}) 24-hour and (\textbf{b}) 72-hour basin-average rainfall depths for East Fork San Jacinto above Cleveland. Both estimates use the complete source water years 2003--2020
and a common Poisson--generalized Pareto tail model with three exceedances per year. RainyDay uses 366 source storms at a candidate rate of $366/18$ storms per year, with 100 realizations of 1,000 simulated years each. This candidate rate is distinct from the tail-exceedance rate. Lines show the Stage IV estimate and the mean RainyDay estimate.
Brown bands show 90\% intervals from resampling observed water years. Lavender bands show 90\% simulated-water-year bootstrap intervals, based on 500 resamples distributed across the 100 realizations. Dotted limits mark the 5th--95th percentile range of the realization estimates.
Both frequency curves draw on the same source period, rather than
an independent held-out record.The RainyDay uncertainty summaries are conditional on the finite
source catalog and simulation model; simulated years do not add independent observed climate exposure.}
\label{figS:rainyday_frequency}
\end{figure}

\begin{figure}[!ht]
\centering
\includegraphics[width=\linewidth]{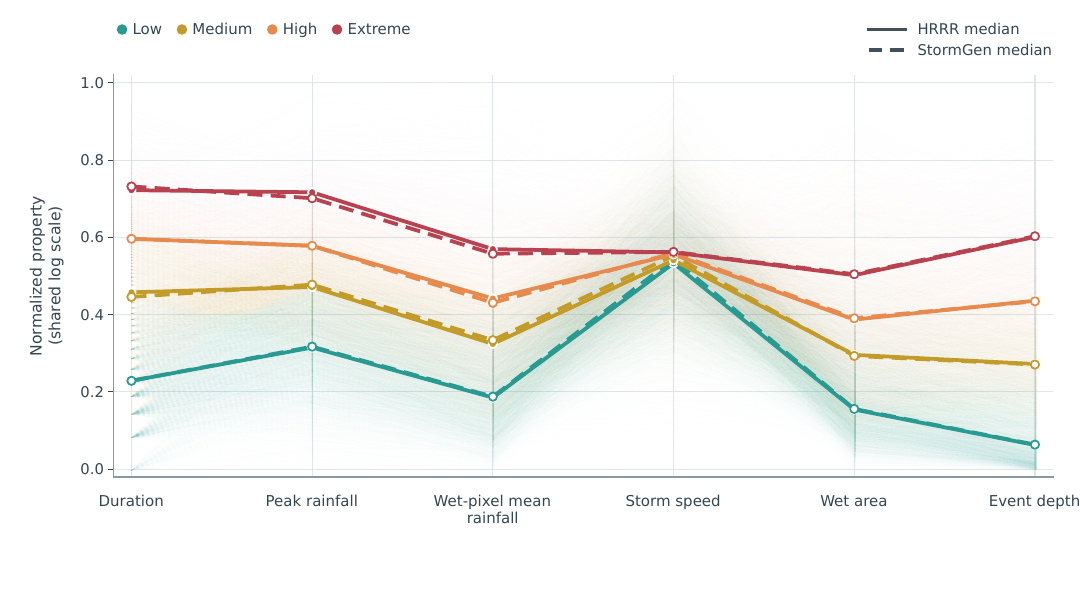}
\caption{\textbf{Storm-property profiles by severity class.}
Faint lines connect six properties of individual catalog entries:
duration, peak rainfall, wet-pixel mean rainfall, storm speed,
wet-area fraction, and event depth.
Colors distinguish low, medium, high, and extreme classes, separated
by shared reference-rarity thresholds at the 50th, 80th, and 95th
percentiles; ties are assigned to the upper class.
Thick solid lines with filled markers show HRRR-labelled class medians,
and thick dashed lines with open markers show StormGen-labelled
class medians.
For property j, the plotted coordinate is
[(1+xj)-aj]/(bj-aj), where aj and bj are the pooled
minimum and maximum of the log-transformed property across both
catalogs. Both catalogs use the same scale without clipping;
constant properties map to zero.
The connecting lines link different properties, not successive times.
Class profiles show coordinatewise medians and need not describe
any single storm.}
\label{figS:storm_properties_profiles}
\end{figure}
\clearpage

\begin{figure}[!ht]
\centering
\includegraphics[width=.7\linewidth]{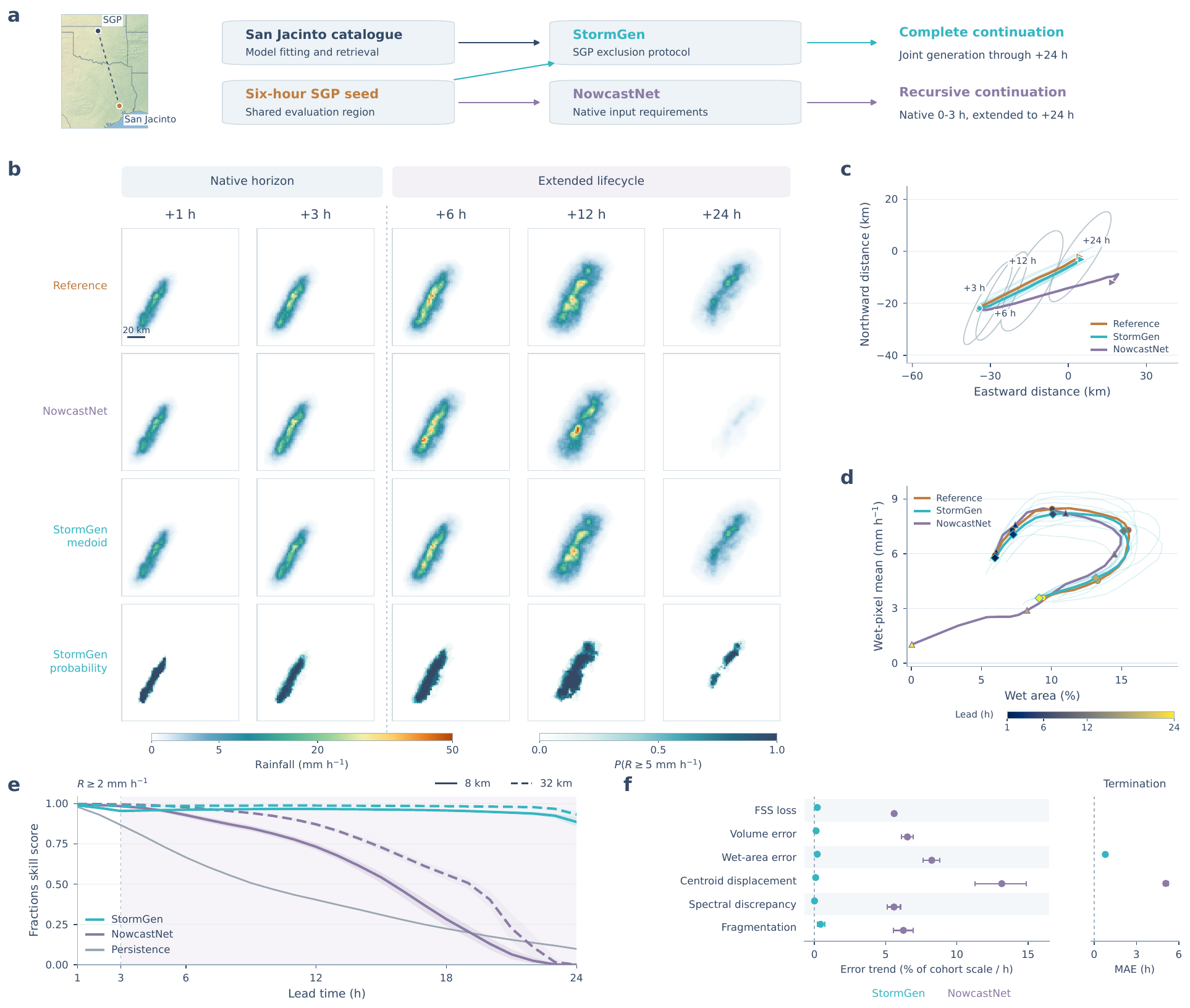}
\caption{\textbf{Illustrative comparison of nowcasting and full-storm continuation.}
The fields and forecasts are constructed examples; their diagnostics illustrate the evaluation framework rather than establish measured model skill.
\textbf{a,} Proposed cross-site evaluation linking the San Jacinto catalogue to six-hour Southern Great Plains (SGP) seeds. Saved history frames illustrate the conditioning input. Parallel timelines contrast joint StormGen generation through +24~h with recursive three-hour NowcastNet rollouts.
\textbf{b,} Rainfall at various hours for the reference, NowcastNet, and a single StormGen ensemble medoid, followed by StormGen exceedance probabilities, $P(R \geq 5~\mathrm{mm\,h^{-1}})$. The medoid minimizes whole-sequence distance to other members. Rainfall panels share a spatial extent and colour scale; probabilities use a separate scale. The divider separates the native nowcasting horizon from the extended continuation.
\textbf{c,} Rainfall-weighted centroid tracks. Thin cyan lines show StormGen members, and the thick cyan line shows their pointwise median. Ellipses depict one-standard-deviation reference rainfall footprints at +3, +6, +12, and +24~h, rather than confidence regions.
\textbf{d,} Lifecycle trajectories relating wet-area percentage to mean rainfall intensity over wet pixels ($R \geq 1~\mathrm{mm\,h^{-1}}$). Marker colour indicates lead time; faint cyan curves show individual StormGen members.
\textbf{e,} Fractions skill score for $R \geq 2~\mathrm{mm\,h^{-1}}$ at 8-km (solid) and 32-km (dashed) neighbourhood scales, with persistence shown at 8~km. The vertical line and shading identify leads beyond the native three-hour horizon. Bands show 95\% paired seed-bootstrap intervals (1,000 resamples; 24 seeds and 12 StormGen members per seed).
\textbf{f,} Post--3~h error trends for six diagnostics, plotted on a common vertical scale as percentages of each metric's pooled 95th-percentile error per hour; positive values indicate worsening. Points show mean trends, and whiskers show 95\% seed-bootstrap intervals. The rightmost comparison reports termination-time mean absolute error on a separate scale, using follow-through to +36~h.}
\label{figS:nowcast}
\end{figure}
\clearpage

% \begin{figure}[!ht]
% \centering
% \includegraphics[width=\linewidth]{figures/figS2_storm_properties_distribution.pdf}
% \caption{\textbf{Individual storm-property paths within severity classes.}
% Each faint line connects the six properties of one catalog entry,
% preserving its across-property associations.
% Solid and dashed paths distinguish HRRR-labelled and StormGen-labelled
% entries, respectively, while color identifies severity class.
% The entries, shared logarithmic scaling, and class boundaries are
% the same as in Fig.~\ref{figS:storm_properties_profiles}.
% Median profiles are omitted to show the spread and overlap of
% individual paths.
% The horizontal axis separates storm properties rather than time,
% and severity classes represent event-percentile groups, not annual
% return periods.}
% \label{figS:distributions}
% \end{figure}
% \clearpage

\end{document}